\documentclass[transmag]{IEEEtran}
\usepackage{latexsym}
\usepackage{graphicx}
\usepackage{amsfonts,amssymb,amsmath}
\usepackage{hyperref}
\def\BibTeX{{\rm B\kern-.05em{\sc i\kern-.025em b}\kern-.08em T\kern-.1667em\lower.7ex\hbox{E}\kern-.125emX}}

\usepackage[usenames,dvipsnames]{color}
\usepackage{tikz}
\usetikzlibrary{shapes,arrows,shadows, positioning}
\usepackage{bm,times}
\usepackage{smartdiagram}

\usepackage{ulem}
\usepackage{amsmath}
\usepackage{amsthm}
\usepackage{amssymb}
\usepackage{bbold}
\usepackage{bm}

\definecolor{gray}{rgb}{0.5,0.5,0.5}
\definecolor{red}{rgb}{0.8,0,0}
\definecolor{dred}{rgb}{0.5,0,0}
\definecolor{blue}{rgb}{0,0.1,1}
\definecolor{dblue}{rgb}{0,0.1,0.6}
\definecolor{cyan}{rgb}{0,0.5,.5}
\definecolor{dcyan}{rgb}{0,0.3,.3}

\definecolor{b}{rgb}{0,0,.8}	
\definecolor{g}{rgb}{0,.6,0}	
\definecolor{n}{rgb}{0,0,0}	
\definecolor{h}{rgb}{0.4,0.2,0.2}	
\definecolor{v}{rgb}{0.2,0.6,0}

\newcommand{\HH}{{\mathcal{H}}}
\newcommand{\II}{{\mathcal{I}}}
\newcommand{\JJ}{{\mathcal{J}}}

\newcommand{\LL}{{\mathcal{L}}}

\newcommand{\PP}{{\mathcal{P}}}

\newcommand{\bsx}{\boldsymbol x}

\newcommand{\bsbeta}{\boldsymbol \beta}

\newcommand{\ov}\overline
\newcommand{\what}{\widehat}
\newcommand{\wtilde}{\widetilde}

\newcommand{\rig}\right
\newcommand{\lef}\left
\newcommand{\nf}\normalfont

\usepackage{subcaption} 
\newcommand{\AD}{\text{AD}}

\begin{document}

\title{
Smoothed Bernstein Online Aggregation \\ for Day-Ahead Electricity Demand Forecasting
}

\author{Florian Ziel
\thanks{Florian Ziel is with 
the House of Energy Markets and Finance, University of Duisburg-Essen, Germany (e-mail: florian.ziel@uni-due.de)}
}

\IEEEtitleabstractindextext{\begin{abstract}
We present a winning method of the IEEE DataPort Competition on Day-Ahead Electricity Demand Forecasting: Post-COVID Paradigm. The day-ahead load forecasting approach is based on online forecast combination of multiple point prediction models. It contains four steps: i) data cleaning and preprocessing, ii) a holiday adjustment procedure, iii) training of individual forecasting models, iv) forecast combination by smoothed Bernstein Online Aggregation (BOA). The approach is flexible and can quickly adopt to new energy system situations as they occurred during and after COVID-19 shutdowns. The pool of individual prediction models ranges from rather simple time series models to sophisticated models like generalized additive models (GAMs) and high-dimensional linear models estimated by lasso. They incorporate autoregressive, calendar and weather effects efficiently. All steps contain novel concepts that contribute to the excellent forecasting performance of the proposed method. This holds particularly for the holiday adjustment procedure and the fully adaptive smoothed BOA approach.

%
%
%
%
%
%
\end{abstract}


}

\maketitle

\section{INTRODUCTION}

\IEEEPARstart{T}{he} COVID-19 pandemic led to 
lockdowns and shutdowns all over the world in 2020 and 2021 to reduce the spread of the corona virus SARS-CoV-2 and the resulting COVID-19 disease.
Obviously, mentioned lockdowns and shutdowns impacted substantially the behaviour the people. Thus, also the consumption of electricity changed dramatically during those periods, \cite{narajewski2020changes}. Electricity load forecasting during lockdowns and shutdown periods is a challenging task, but even months afterwards the forecasting task is still complicated. One reason is that is not obvious which of the changed behavioral pattern during the lockdown observed in many countries (e.g. increased remote work, getting up later) will persist months and years after the lockdown. Another problematic aspect is the disruption of annual seasonalities during the lockdown periods. 

\begin{figure}
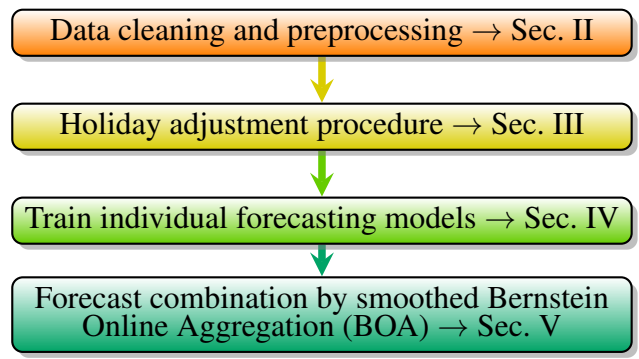

\begin{center}
\smartdiagramset{border color=black, text width=8cm, 
set color list={orange!40!orange,orange!40!lime,orange!40!green,blue!40!green},
back arrow disabled=true, module y sep= 1.25cm, module minimum height=0.5cm}
\smartdiagram[flow diagram]{
\large
Data cleaning and preprocessing $\rightarrow$ Sec. \ref{sec_data}, 
\large 
Holiday adjustment procedure $\rightarrow$ Sec. \ref{sec_holiday}, 
\large 
Train individual forecasting models $\rightarrow$ Sec. \ref{sec_forecasters}
,
\large Forecast combination by smoothed Bernstein Online Aggregation (BOA) $\rightarrow$ Sec. \ref{sec_boa} }
\end{center}
\caption{Structure of forecasting approach used for the forecasting competition.}
\label{fig_model_flow}
\end{figure}

The IEEE DataPort Competition \emph{Day-Ahead Electricity Demand Forecasting: Post-COVID Paradigm} focuses on Post-COVID aspects in electricity load forecasting \cite{67vy-bs34-20}. The day-ahead load forecasting competition was based on real data and run over a test period of 30 days.
This manuscript describes one of the winning method that scored 3rd in the competition\footnote{According to significance test conducted by the organizers, the top 3 positions where not significantly different from each other.}.
The prediction approach is based on smoothed 
Bernstein Online Aggregation (BOA) applied on individual load forecasting models. The full model flow is depicted in Figure 
\ref{fig_model_flow}.

The manuscript is organized as follows. 
First we introduce the data set and the forecasting task in more detail and discuss inital data preprocessing steps. Afterwards, we 
explains a holiday-adjustment procedure to deal adequately with holidays in the data. Section \ref{sec_forecasters} introduces multiple individual forecasting models that are mainly (high-dimensional) statistical forecasting models that are sometimes referred as experts or base learners. Then, we descripe the expert aggregation procedure BOA with a smoothing extention.
We conclude with some final remarks.

%
%
%
%
%
%

\section{Data and preprocessing} \label{sec_data}
\begin{figure*}[htb!]
\centerline{\includegraphics[width=0.999\textwidth]{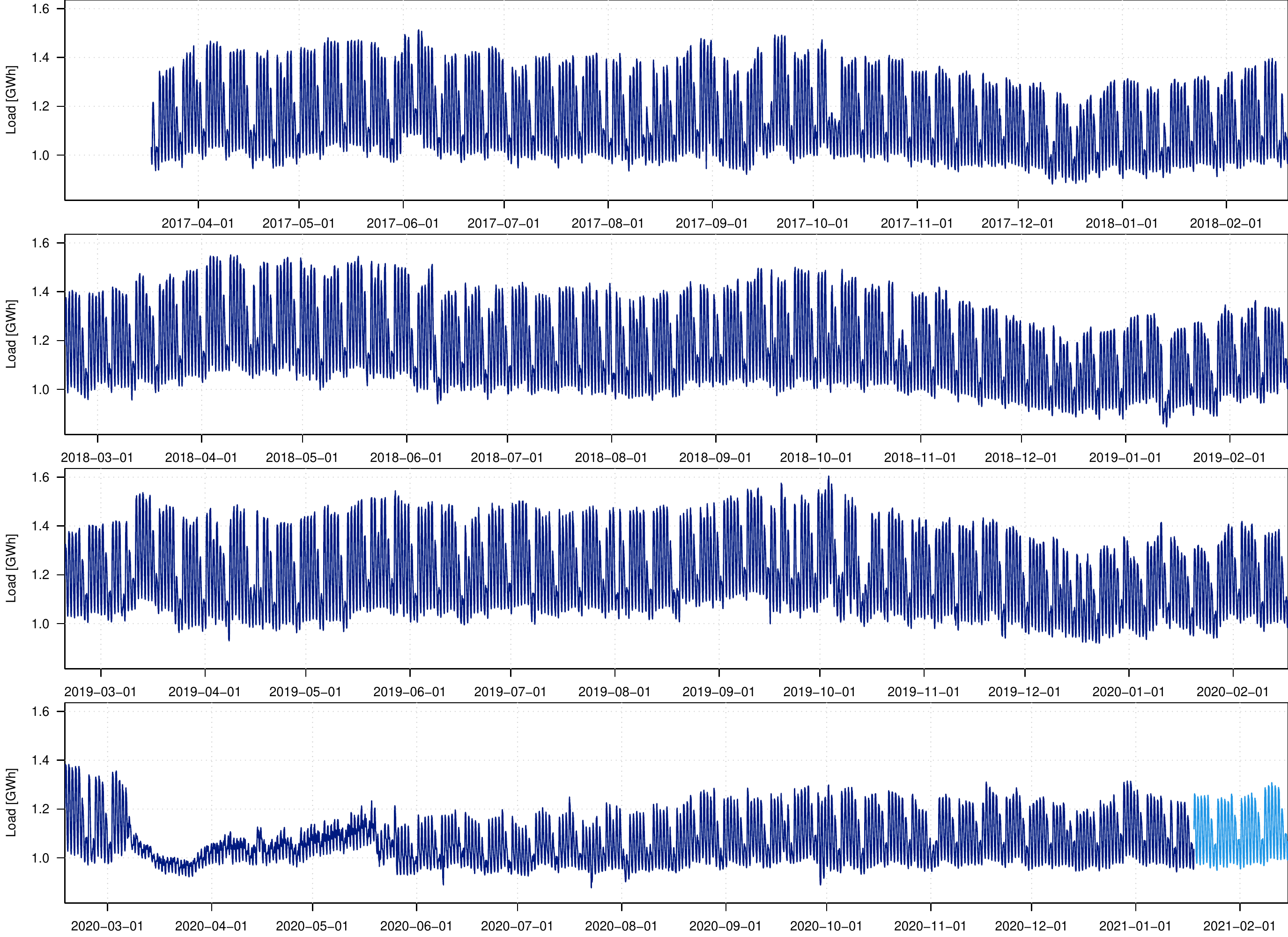}}
\caption{Available load data for day-ahead for the competition. The test data is highlighted in light blue. \label{fig1}}
\end{figure*}

The load forecasting competition contains initially hourly load data from 2017-03-18 00:00 to 2021-01-17 07:00, visualized in Figure \ref{fig1}. According the organizers the load data corresponds to one city, but the origin of the load data to predict was disclosed.

\begin{figure*}[htb!]
\centerline{\includegraphics[width=0.999\textwidth]{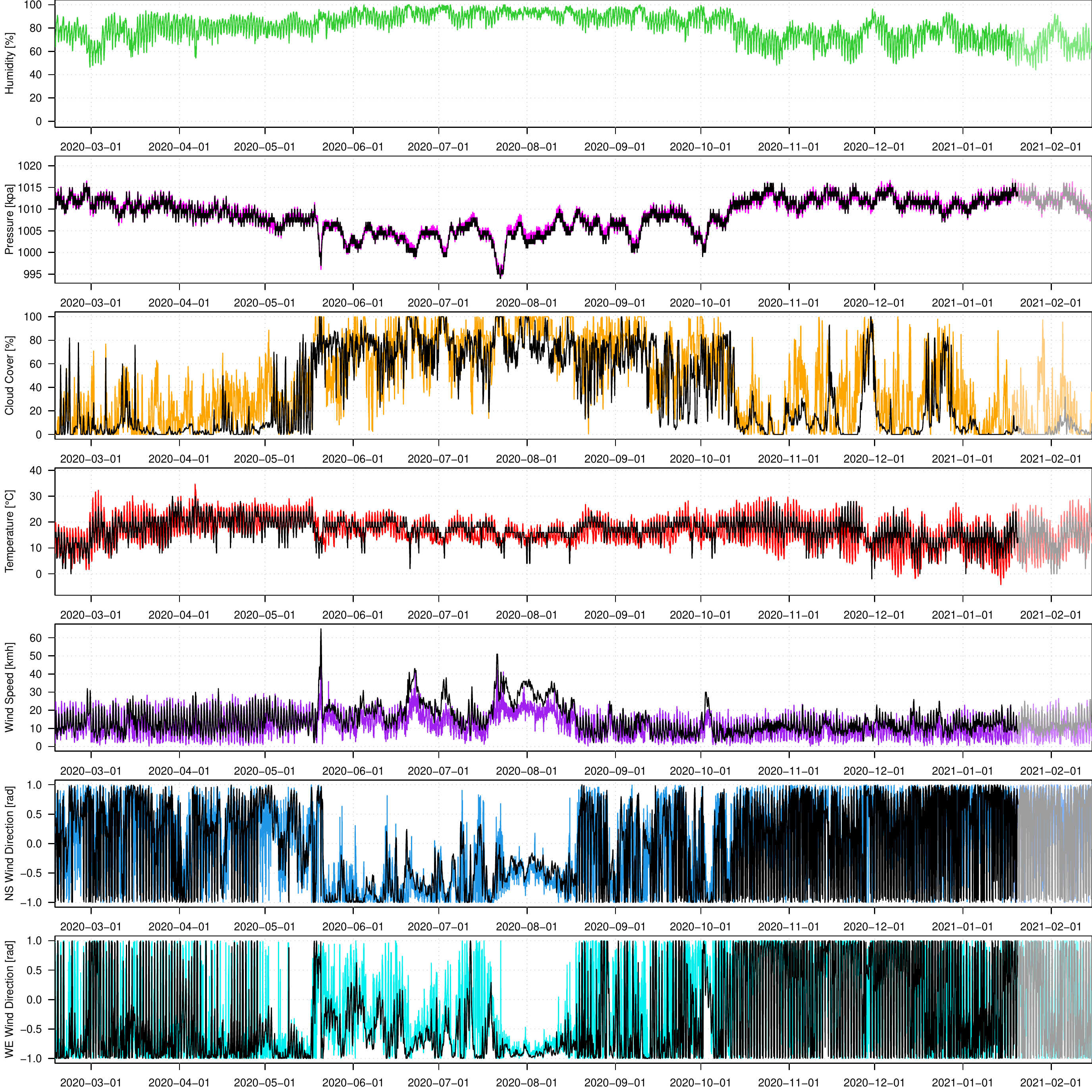}}
\caption{External input weather data available for the competition for the last year. Colored data correspond to actuals and black ones to day-ahead forecasts. The test data is indicated by lighter colors.
\label{fig2}}
\end{figure*}

The daily forecasting task is to predict the next days hourly load, which corresponds to forecast 24 values 17 to 40 hours ahead.
Thus, the first forecasting task was aiming for the hourly load for 
2021-01-18 from 00:00 to 23:00. The second task was to predict the load on 2021-01-19. This rolling forecasting procedure was continued over 30 days in the competition.
In the bottom chart of Figure \eqref{fig1} you see clearly the structural break due to the COVID-19 lockdown in March 2020. The overall load level dropped and the weekly profile got disturbed dramatically. In the proceeding months we observe some slowly increasing recovery of the electricity consumption. However, even in 2021 we observe that especially the peak hours have a lower load level than the previous years.

Next to the actual load data, also weather input data was provided. This was actual data on humidity, pressure, cloud cover, temperature, wind speed such as day-ahead forecasts of all meteorologic features except humidity were provided, Figure \ref{fig2} for last years data. The day-ahead weather forecasts were in fact 48-hours ahead forecast. Thus, for the first day, weather forecasts data up to 2021-01-19 07:00 was provided.
During the competition the actual load and weather data, and the weather forecast data for the next 24 hours were released, leading to a typical rolling forecasting study design.

The weather data contained some obvious reporting problems which were cleaned using linear interpolation and the \texttt{R}-package \texttt{tsrobprep}, see \cite{narajewski2021tsrobprep, tsrobpreppackage}.
Afterwards, we transformed the wind direction data to the north-south (NS) and east-west (EW) component by evaluating the cosine and sine of the wind direction data.
Thus, Figure \ref{fig2} shows the cleaned data for the available weather forecasts and actuals. For further analysis, we extend the weather data input space by adding rolling daily means of all weather inputs.

The evaluation metric is the mean absolute error (MAE) which corresponds to point forecasting. More precisely, median forecasts are required to minimize the MAE, see \cite{gneiting2011making}.

\section{Holiday adjustment procedure} \label{sec_holiday}
As the origin of the data was disclosed and no holiday calendar was provided a specific solution for dealing with holidays is required.
Handling holidays adequately is an important task and may improve the forecasting accuracy substantially even for the non-holidays, see e.g. \cite{ziel2018modeling}.

\begin{figure*}[htb!]
\centerline{\includegraphics[width=\textwidth]{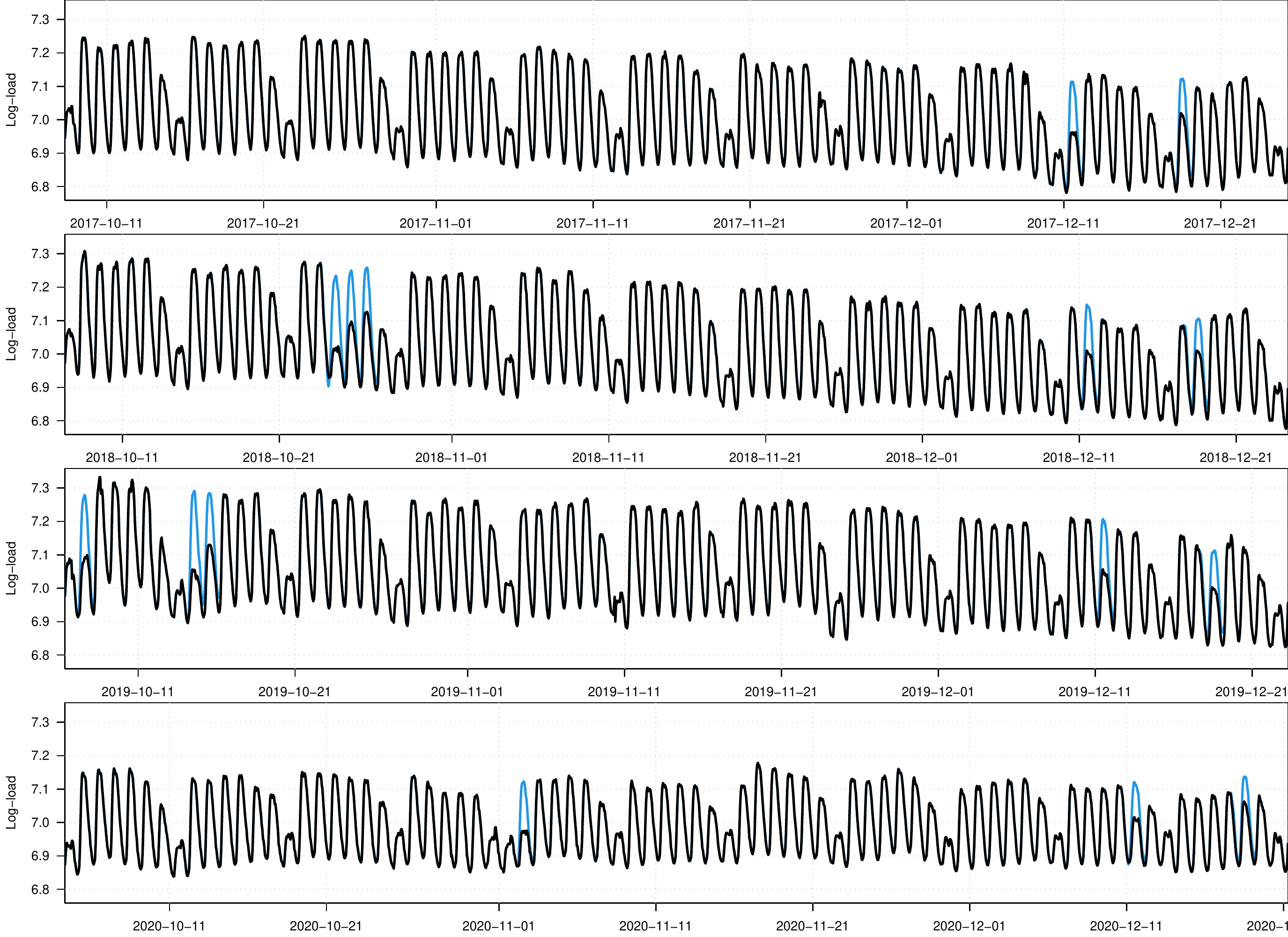}}
\caption{Log-load data in October to December in considered years (black) with holiday adjustment of the proposed procedure (blue). \label{fig_holiday}}
\end{figure*}

By eyeballing, it is easy to spot some obvious date-based public holidays in the data (12Jan, 17Apr, 1Aug, 18Sep, 11Dec, 18Dec). But there are also a couple days which behave like holidays but the pattern of occurrence seems to be different.
We consider a holiday adjustment procedure to
take into account the holiday impact appropriately.
%
The procedure is based on a high-dimensional time series model, similarly used in the GEFCom2014 (Global Energy Forecasting Competition 2014), see \cite{ziel2016lasso}.
The result of the considered procedure is illustrated for the period from October to December in Figure \ref{fig_holiday}.

To introduce the holiday adjustment procedure formally, we require some notations. Denote $\ell_t =\log(L_t)$ the logarithm of the load $L_t$ at time point $t$.
Let $T$ be the number of observations that is currently available for model training.
The considered model is a high-dimensional linear model for $\ell_t$ containing the following components in the input matrix:
\begin{enumerate}
 \item[i)] lagged log-load $\ell_{t+k}$ values, for $k\in \II_{\text{pos}}\cup \II_{\text{neg}}$ where  $\II_{\text{pos}} = \{168,169,\ldots,510\}$ and $\II_{\text{neg}}= \{-168,-169,\ldots,-510\}$
 \item[ii)] $p$-quantile ReLU-transformed (Rectified Linear Unit transformed) of all available weather data on quantile grid
 of probabilities $\PP = \{0,0.1,\ldots,.9\}$. In detail we compute 
$x^{p\text{-ReLU}}_t = \max\{x_t - q_p(\bsx), 0\} $ 
with $q_p(\bsx)$ for $p\in \PP$ as $p$-quantile of $\bsx$ for weather input feature $\bsx = (x_1,\ldots, x_T)'$.
 \item[iii)] All weather data interactions, i.e. $x_{x_t,y_t,t}^{\text{inter} } = x_t y_t$ for inputs $x_t$ and $y_t$.
 \item[iv)] Daily and weekly deterministic effects.
 The daily and weekly effects are modeled by standard and cumulative dummies:
 \begin{align}
  x_{k,t}^{\text{day}} &= \mathbb{1}\{\text{HoD}_k(t)=k\} \text{ for } k\in\{1,\ldots, 24\} \label{eq_dummy_day}\\
  x_{k,t}^{\text{cday}} &= \mathbb{1}\{\text{HoD}_k(t)\leq k\} \text{ for } k\in\{1,\ldots, 24\} \label{eq_dummy_cday} \\
  x_{k,t}^{\text{week}} &= \mathbb{1}\{\text{HoW}_k(t)=k\} \text{ for } k\in\{1,\ldots, 168\} \label{eq_dummy_week} \\
  x_{k,t}^{\text{cweek}} &= \mathbb{1}\{\text{HoW}_k(t)\leq k\} \text{ for } k\in\{1,\ldots, 168\} \label{eq_dummy_cweek}
 \end{align}
where $ \text{HoD}_k(t)$ and $ \text{HoW}_k(t)$ are the hour-of-the-day and hour-of-the-week dummies.
 \item[v)] Annual deterministic effects described by 
periodic cubic B-splines with annual periodicities ($A=24\times 365.24$ hours). Precisely, we consider 12 basis functions on a equidistant grid on $[0, A)$. For more details on periodic cubic B-splines in energy forecasting see \cite{ziel2016forecasting}.
\item[vi)] Impact-adjusted holiday dummies on days which were identified in advance as potential holidays. 
\end{enumerate}
The lagged log load in i) describes the autoregressive impact on a specific day for the surrounding 3 weeks of information without using nearby information of the surrounding week, to exclude any impact from bridging effects. 

Note that ReLU-transformed weather input in ii) is relevant to capture non-linear weather impacts. However, for $p=0$ the linear effect is modelled.
 Component iii) is motivated from the second order Taylor approximation. Considering all weather data interactions allows us to capture relevant non-linear information.
In fact, components ii) and iii) may be regarded as a manual application of the kernel trick to the input data to enlarge the feature space.

Further, in iv) the standard dummies with '='-sign in the definition (see \eqref{eq_dummy_day} and \eqref{eq_dummy_week}), have the job of detecting demand effects that happen only at the day or week period (e.g. if the load is high only at a certain hour of the day). In contrast, the cumulative dummies (see \eqref{eq_dummy_cday} and \eqref{eq_dummy_cweek}) have the purpose to describe effects that persists over multiple hours in the day or week period.

The component vi) models the holiday effect and is crucial for the holiday adjustment procedure. Its design corresponds to the holiday modeling approach used in see \cite{ziel2016lasso}. However, next to the impact multiplication also a scaling of the impact. Precisely it is scaled by the difference of rolling quantiles at probabilities 90\% and 37\% of the previous week. The idea is that the upper quantile is an estimate standard activity in a working week and the lower quantile and estimate for the Sunday peak.
This adjustment procedure is required to deal with the strong structural breaks during the COVID-19 shutdown. This, effect can be seen in
Figure \ref{fig_holiday} as well. We observe that the absolute holiday impact of 
11th December is smaller in 2020 than the years before.

The model for the log-load $\ell_t$ with all inputs i) to vi) is estimated using lasso (least absolute shrinkage and selection operator) on scaled input data. The tuning parameter is chosen by minimizing the Bayesian information criterion (BIC), see e.g. \cite{zou2007degrees}. Now, we take the fitted parameter vector $\what{\bsbeta}$ and set all estimated parameters which correspond to the holiday impacts vi) to zero, to receive $\what{\bsbeta}^{\text{hldadj}}$. The fitted values with respect to $\what{\bsbeta}^{\text{hldadj}}$ is the holiday-adjusted log-load time series $\tilde{\ell}_t$,
as illustrated in Figure \ref{fig_holiday} in blue.

Note that for the inital and final three weeks (exactly 510 hours as the maximum in $\II_{\text{pos}}$) the procedure can not be applied as $\ell_{t+k}$ is not available all the time. Therefore, we train for the inital three weeks the same model without $\II_{\text{neg}}$
and for the last three weeks the model without $\II_{\text{pos}}$.

The complete lasso training procedure including tuning parameter selection on the full data set takes around half a minute on the authors laptop using \texttt{glmnet} of \texttt{R} on a single core. However, it is important to use sparse matrix support to reduce computation time. 

%
%

 \section{Training of individual forecasting models} \label{sec_forecasters}
 
 Given the holiday adjusted log-load $\tilde{l}_t$ and the resulting 
 holiday adjusted load $\wtilde{L}_t$ we train many forecasting models to create a big pool of forecasters (or experts).
 The considered models range from simple time series model more advanced statistical learning procedures.
 Also several non-linear models
 gradient boosting machines (GBM) (using the \texttt{R} packages \texttt{gbm} and \texttt{lightgbm}) and neural networks (using the \texttt{R} packages \texttt{nnet} and \texttt{keras}) were tested. 
 But the forecasting accuracy was rather low and they did not improve the forecasting performance in the forecasting combination method described in Section \ref{sec_boa}. The reason might be that the major impacts are linear, esp. autoregressive and seasonal effects.
 
The considered models, can be categorised into four types.
 This is 
 \begin{enumerate}
          \item[A)] STL-decomposed exponential smoothing 
          $\rightarrow$ Sec. \ref{subsec_stl}
          \item[B)] AR($p$) models
                    $\rightarrow$ Sec. \ref{subsec_arp}

          \item[C)] Generalized additive models (GAMs)
                    $\rightarrow$ Sec. \ref{subsec_gam}

          \item[D)] Lasso estimated high-dimensional linear regression models
                    $\rightarrow$ Sec. \ref{subsec_lasso}

         \end{enumerate}
%
The lasso type model had best individual prediction accuracy. 
Further, all models are applied to the holiday adjusted load time series and the holiday adjusted log-load $\tilde{l}_t$ and the 
 holiday adjusted load $\wtilde{L}_t$. For convenience, we introduce the notation $Y_t \in \{\tilde{l}_t,\wtilde{L}_t\} $. When considering a log-load model, the exponential function is applied to the point forecasts $\what{\wtilde{\ell}}_{T+h}$ for the forecasting horizon $h\in \HH= \{h_{\min},\ldots,h_{\max}\}=\{17, 18,\ldots, 40\}$ to predict the load at $T+h$.
 
All models
were estimated using a calibration window size of $C\in \{
28,   56,   77,  119,  210,  393,  758, 1123\}$ days minus 16 hours (as the last available data point was at 8am). 
The general idea behind this is quite simple, models with short calibration windows (e.g. 4, 8, 12 weeks) shall adjust better to more recent data, models with larger windows have more data to learn better about rare event like the annual effects.
Moreover, several forecasting studies in energy forecasting have shown that combining short and long calibration windows, may lead to substantial gain in forecasting performance, see e.g.  
\cite{hubicka2018note, maciejowska2020pca}.

The described forecasting procedure was applied in a rolling forecasting study to all days starting from 1st June 2020 as first day to predict. This date was chosen by manual inspection the historic data, as the hard COVID-19 shutdown effects seem to be vanished.

\subsection{STL decomposition with Exponential Smoothing}
\label{subsec_stl}

This approach applies first an STL decomposition on $Y_t$. STL acronym represents the decomposition into to trend, seasonal and remainder components by loess (locally weighted scatterplot smoothing).

On the remainder component an additive exponential smoothing model is fitted. This is done using the \texttt{stlf} function of the \texttt{forecast} package in \texttt{R}, \cite{forecastpackage}. The seasonality of the time series are set to 168. Forecasting is done recursively for forecasting horizon up to $h_{\max}$, and report $h_{\min},\ldots,h_{\max}$.


\subsection{AR(p) time series model}
\label{subsec_arp}

Here, $Y_t$ is modeled by a simple autoregressive process ($AR(p)$) where $p$, sometimes used in energy forecasting \cite{ziel2016iteratively, steinert2019short}. The only tuning parameter $p$ is selected by minimizing the Akaike information criterion (AIC) with $p_{\max} = 24\times22=528$ (3 weeks plus 1 day). This done using the R function \texttt{ar} of the \emph{stats} package in \texttt{R}, see \cite{statspackage}. Again, the forecasting is done recursively
to $h_{\max}$, and report $h_{\min},\ldots,h_{\max}$.

\subsection{Generalised additive models (GAMs)}
\label{subsec_gam}

This procedure utilized generalised additive models which are popular in load forecasting, see e.g. the winning method of the Global Energy Forecasting Competition 2014 in the load track \cite{gaillard2016additive}.

In fact we consider 2 separate GAM model designs due to the limited accessibility of the $Y_{t-24}$ for forecasting horizons $h\in \HH$. 
For hour the first 8 horizons $h \in \{17,\ldots,24\}$ the GAM model is
\begin{align*}
Y_{t} \sim & \underbrace{ \sum_{k\in \{24,168\}} s(Y_{t-k}) + \sum_{k\in \JJ} Y_{t-k}}_{\text{autoregressive effects (non-linear + linear)}} + \underbrace{u(\text{hour},\text{week})}_{\text{weekly profile term}}\\
 		 & +\underbrace{ s(f_{\text{temp}})+s(f_{\text{temp}},\text{week}) + s(f_{\text{cc}}) +s(f_{\text{cc}},\text{week}) }_{\text{non-linear temperature and cloud cover effects depending on weekday}} \\ 
 		 &+ \underbrace{ s(f_{\text{temp}},f_{\text{cc}})}_{\text{temperature and cloud cover interaction}} \\ 
 		 &+ \underbrace{  s(f_{\text{pres}}) + s(f_{\text{wind}}) + s(f_{\text{dircos}})+ s(f_{\text{dirsin}})}_{\text{non-linear effects from pressure, wind speed and direction}} 
\end{align*}
for index set $\JJ = \JJ_{\text{short}} \cup \JJ_{\text{long}}$ 
with $\JJ_{\text{short}}= 24\cdot \{2,3,8,14,21,28,35,42\}$ and $\JJ_{\text{long}}= 24\cdot \{350,357,364 ,371 ,378 ,385\}$.
Here, $s$ represents regression smoothing terms and $u$ tensor products, $f_*$ represent the forecasts of the daily rolling averages of the meteorologic components. The inputs hour and week take values 1,\ldots, 24 and 1,\ldots,7 depending on the corresponding time $t$.
For horizons $h> 24$, the term $s(Y_{t-24})$ in the model is dropped.

The autoregressive terms capture the dependency structure of the past for the corresponding hour. Note that the yesterdays load $Y_{24}$ and previous weeks load $Y_{168}$ is regarded as very important and therefor non-linear effects are considered. Preliminary analysis showed that the weather variables temperature and cloud cover are more relevant to explain the load behavior than other weather variables. There, we included next plain non-linear effects on each individual variable which potentially varies over the week also interaction effects. The remaining weather variables enter with non-linear smoothing effects.

The models are trained by considering only the data of the corresponding target hours. Obviously, the forecasting is done directly. The implementation is done using the \texttt{gam} function
of the \texttt{R}-package \texttt{mgcv}, see \cite{gampackage}.

\subsection{Lasso based high-dimensional regression models}
\label{subsec_lasso}

The lasso based models are very similar to the model used for 
the holiday adjustment in Section \ref{subsec_lasso}.
Therefore, we only highlight the differences which concerns 
the autoregressive design and details on the estimation procedure.

The high-dimensional linear models are trained for each forecasting horizons $h\in \HH$ separately.
Additionally, the lag sets $\II_h$ are adjusted to 
$\II_h= \II_{h,\text{day}}\cup \II_{h,\text{week}} \cup \II_{h,\text{year}}$
with 
$\II_{h,\text{day}}= \cdot \{h, \ldots, 24\ldots 15+h\} - h$, 
$\II_{h,\text{week}}= 24\cdot\{21,28,\ldots, 56\} - h$
and 
$\II_{h,\text{year}}= 24\cdot\{350,357,364,371\} - h$, 
for $h\in \HH$ to incorporate daily, weekly and annual autoregressive effects.
%
%
The high-dimensional regression model is trained by lasso on an exponential tuning parameter grid of size 20. In detail the grid for the regularization parameter $\alpha$ is 
$2^\LL $ where $\LL$ is an equidistant grid form $6$ to $-1$ of size 20.

		 

\section{Forecast combination by smoothed Bernstein Online Aggregation (BOA)} \label{sec_boa}

After creating all forecasting models as described in Section 
\ref{sec_forecasters}, an online aggregation procedure is used to combine the forecasts. 
The combination method is based on an extension of the fully adaptive Bernstein Online Aggregation (BOA) procedure, see \cite{wintenberger2017optimal}. The BOA is extended by a smoothing component and is implemented in the \texttt{R} package \texttt{profoc} \cite{profoc_package}.
It is similarly as used in \cite{berrisch2021crps} for CRPS learning.

\subsection{Formal description of the algorithm}
To introduce the smoothed BOA formally, we require some further notations.
Denote $\what{L}_{d,h,k}$ the available load forecasts for forecast issue day $d$, prediction horizon $h$ and forecasting model $k$.
If current forecast is for day $d$, then we are looking 
for optimal combination weights $w_{d,h,k}$. This is used to combine the predictions linearly so that
\begin{align}
\wtilde{L}_{d,h} = \sum_{k} w_{d,h,k} \what{L}_{d,h,k} 
\end{align}
is the forecast aggregation to report. Moreover, denote
$\AD(x, y) = |y-x|$
the absolute deviation (also known as $\ell_1$-loss) which is a strictly proper score for median predictions and MAE-minimization, see \cite{gneiting2011making}. Additionally, let
$\AD^{\nabla}(x,y) = \AD'(\wtilde{L}_{d,h},y) x$ where $\AD'$ is the (sub)gradient of $\AD$ with respect to $x$ evaluated at forecast combination $\wtilde{L}_{d,h}$. We require $\AD^{\nabla}$ to apply the so called gradient trick to enable optimal convergence rates in the BOA, see \cite{wintenberger2017optimal, berrisch2021crps}.

The smoothed fully adaptive BOA with gradient trick and forgetting has the five update steps. In every update step we update the instantaneous regret $r_{d,h,k}$, the range $E_{d,h,k}$, the learning rate $\eta_{d,h,k}$, the regret $R_{d,h,k}$, and the combination weights $w_{d,h,k}$ for forecasting horizon $h$ and forecaster $k$: 
    \begin{align}
        r_{d,h,k} =& \, \AD^{\nabla}(\wtilde{L}_{d,h},L_t) - \AD^{\nabla}(\widehat{L}_{d,h,k},L_t)\\
        E_{d,h,k} =&  \max(E_{d-1,h,k}, |r_{d,h,k}|)  \\
        \eta_{d,h,k} =&  \min\left(\frac{E_{d,h,k}}{2}, \sqrt{ \frac{\log(K)}{ \sum_{i=1}^t r^2_{i, k}} }\right) \\
        R_{d,h,k} =& \, R_{t-1,k} + r_{d,h,k}  \left(  \eta_{d,h,k}  r_{d,h,k} -1 \right) /2 \nonumber \\ & + E_{d,h,k}  \mathbb{1}\{-2\eta_{d,h,k}r_{d,h,k} > 1 \} \\
        w_{d,h,k} =& \, \frac { \eta_{d,h,k} \exp \left(-  \eta_{d,h,k}  R_{d,h,k} \right) w_{0,h,k} }{   \frac{1}{K}  \sum_{k = 1}^K  \eta_{d,h,k} \exp \left( - \eta_{d,h,k}   R_{d,h,k}\right) } \label{eq_boa_w}
    \end{align}
with inital values $w_{0,h,k}=1/K$, $R_{0,h,k} = 0$ and $E_{0,h,k}=0$.

As it can be seen in equation \eqref{eq_boa_w} the 
BOA considers an exponential updating schema as the popular 
exponential weighted averaging (EWA), see \cite{cesa2006prediction}.
The BOA will lead always to a convex combination of the forecasters, as the EWA.
Further, is well known that the EWA in combination with the 
gradient trick can achieve optimal convergence rates, if the considered updating loss is exp-concave, see \cite{cesa2006prediction}. Unfortunately, the required absolute deviation $\AD$ is not exp-concave.
Therefore, the BOA uses a second order refinement in the weight update to achieve better convergence rates under weaker regularity conditions on the considered loss.
In fact, the mentioned gradient trick and the second order refinement allow the BOA to achieve almost optimal convergence rates for the selection problem and convex aggregation problem. 
\cite{wintenberger2017optimal} and  \cite{gaillard2018efficient} prove that the BOA considered for absolute deviation loss 
has almost linear convergence with respect to the prediction performance of the best individual expert and a almost (standard) square root convergence with respect to the optimal convex combination. 
Both convergence rates are only almost optimal as there is an additional $\log(\log)$ term in both convergence rates which is due to the online calibration of the learning rate.

Now, we motivate the smoothing extension of the BOA:
The described BOA algorithm applies the forecast combination to each target hour $h$ individually. However, it could be a reasonable assumption that the weights $w_{d,h,k}$ are constant across all $h\in\HH$. This restriction reduces the estimation risk in the algorithm for sacrificing theoretical optimality. Hence, we want to find solution between those two extreme situations which finds the optimal trade-off. Therefore, we are considering smoothing splines, applied to the weights $w_{d,h,k}$.
As suggested by \cite{berrisch2021crps} we consider cubic P-splines on an equidistant grid of knots of size 24. 
The smoothed weights $\wtilde{w}_{d,h,k}$ are computed by
\begin{align}
\wtilde{w}_{d,h,k} = B(B'B+ \lambda D'D)^{-1} B' w_{d,h,k} 
\end{align}
where $\lambda\geq 0 $ is a smoothing parameter, $B$ is the matrix of cubic B-splines and $D$ is the difference matrix where the difference operator is applied to the identity. Note that we difference only once, as this implies smoothing towards a constant function if $\lambda\to \infty$, see \cite{berrisch2021crps}. The tuning parameter $\lambda$ has to be determined.

\subsection{Application, parameter tuning and forecasting results}

As explained in the introduction the competition was conducted in a rolling window framework and maps realistic settings. However, for illustration purpose, we concentrate one forecasting task, this is to forecast the 1st February 2021 from 0:00 to 23:00 where the last available observation is on 31st January 2021 7:00.

We decided to utilize a stepwise forward approach to determine which forecasts to combine using the BOA. 
Therefore, we consider a burn-in period of 30 days (to allow local convergence of the BOA) and keep the last 60 days of available data for calibration. The final number of models $M$ to combine was determined by evaluating the MAE of the $M_{\max}=40$ combination procedures on the calibration data set. 
The results for the validation MAE across all forecasting horizons are shown in Figure \eqref{fig_boa_lambda}. Additionally, we label the selected models for the optimal number of models to combine, which is 5 in this situation.
\begin{figure}[htb!]
\centerline{\includegraphics[width=.499\textwidth]{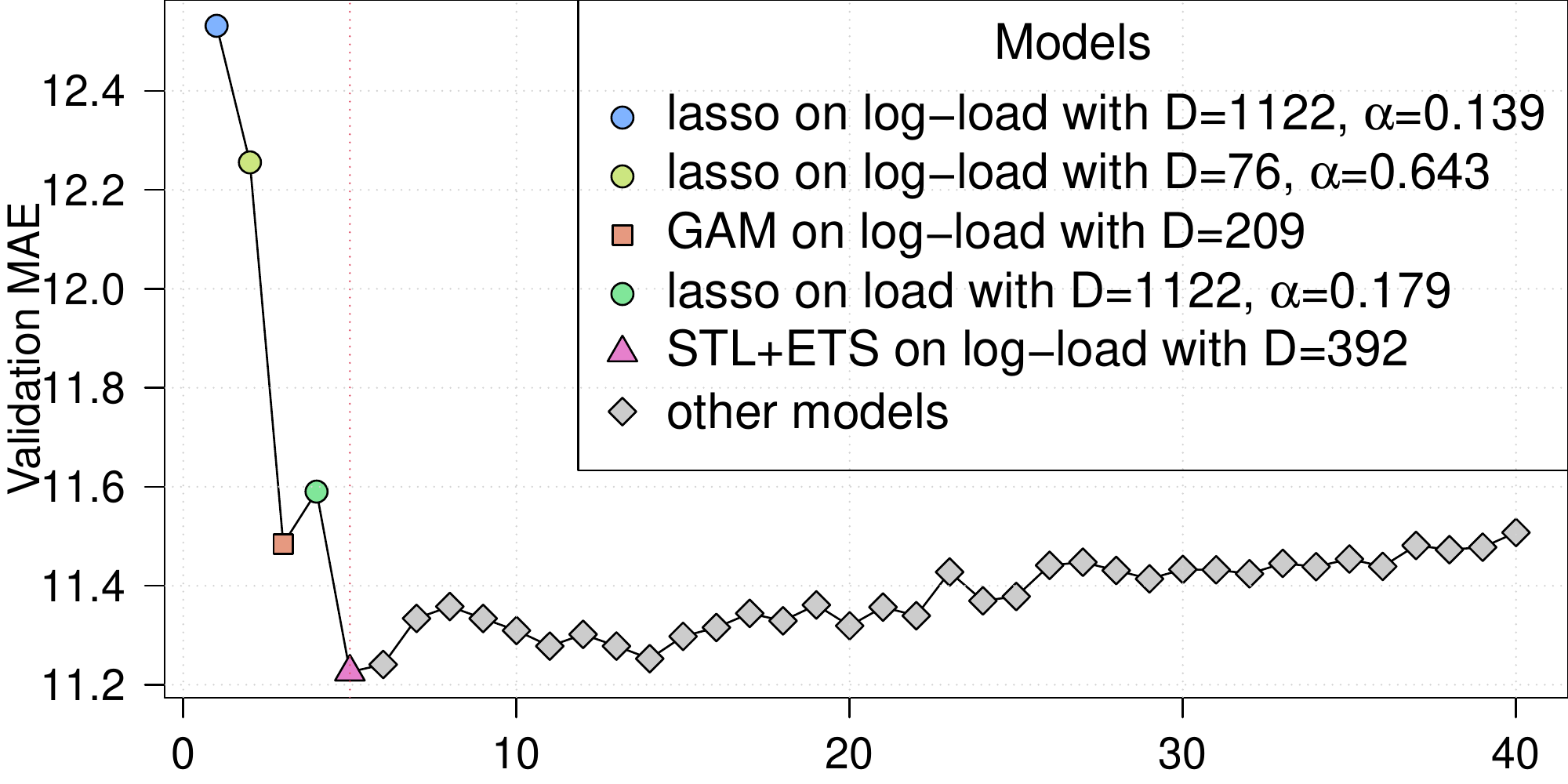}}
\caption{MAE on the validation data set with highlighted optimal number of forecasting models. For the 5 optimal models we show the corresponding calibration window length $D$ and tuning parameter $\lambda$.  }
\label{fig_boa_lambda}
\end{figure}
We observe that especially the first few models contribute substantially to the MAE reduction which is about 10\% compared to the best individual model. It is interesting to see that the selected 5 models are quite diverse. Those are three lasso based models, a GAM model and an STL+ETS model. From the selected lasso models, two use a long history of about 3 years of data and one just a very short history of about 3 months. Also the GAM model considers a relatively short history of 7 months.

After selecting the forecasters to combine we run a BOA algorithm on an exponential $\lambda$-grid. We choose always the $\lambda$-value which performs best in the past to predict the next day. More precisely, we chose the $\lambda$-value so that the exponentially discounted MAE with a forgetting parameter $\rho=0.01$ is minimized. Note that this forget corresponds to an effective sample size of $1/\rho$ which is 100, so about 3 months. 
Figure \eqref{fig_boa_lambda} shows the results for the selected values for the smoothing parameter $\lambda$ on the considered training and validation set.

\begin{figure}[htb!]
\centerline{\includegraphics[width=.499\textwidth]{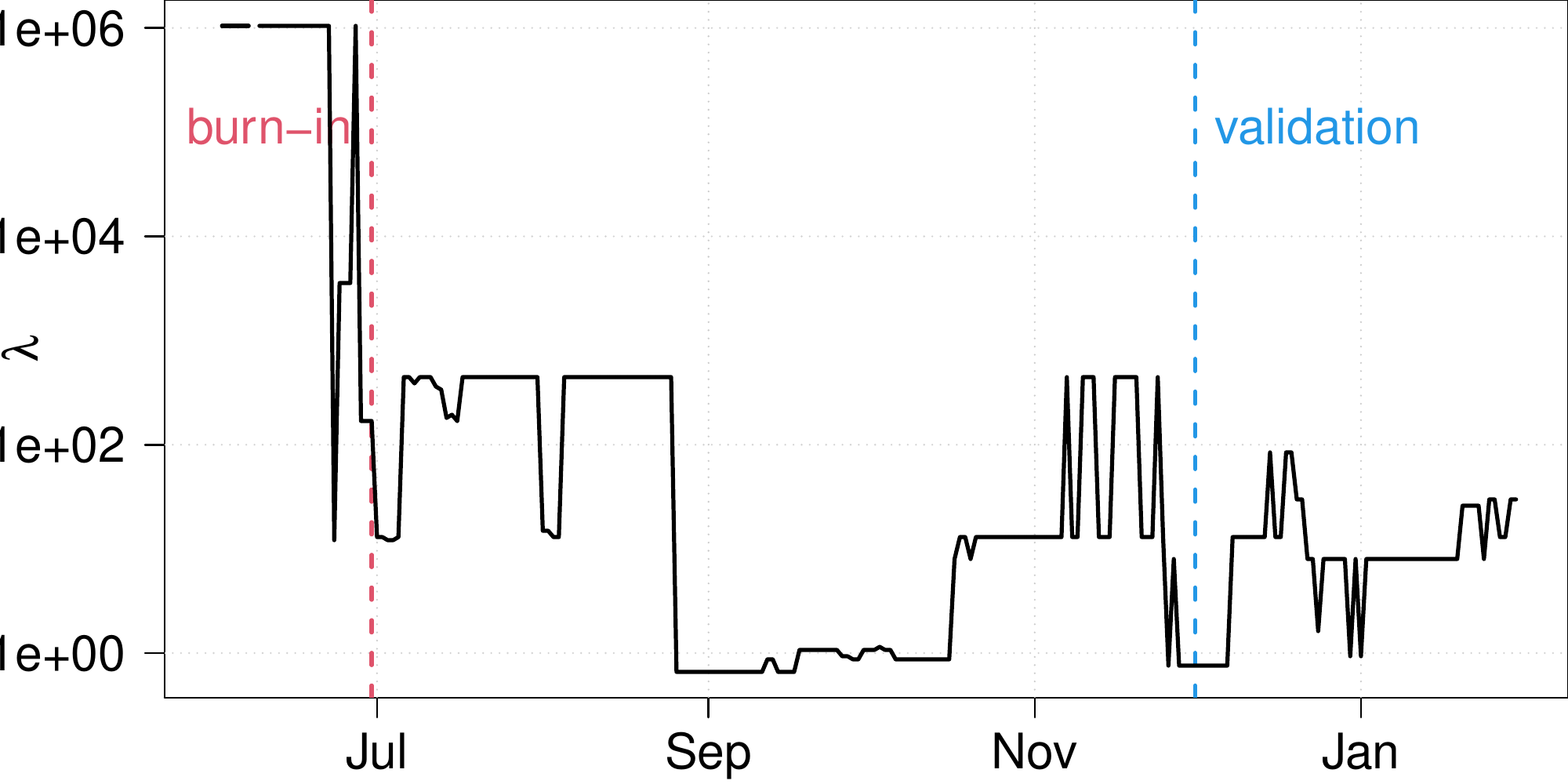}}
\caption{Selected $\lambda$ on the test and validation set with highlighted burn-in and validation period. }
\label{fig_boa_lambda}
\end{figure}
We observe that the selected smoothing parameter clearly varies over time. It is also interesting to see that in the burn-in phase very high $\lambda$ values where selected. This correspond to a conservative selection with low estimation risk. This selection is plausible, as the amount of information to evaluate is low in the burn-in period.

\begin{figure*}[htb!]
\centerline{\includegraphics[width=.999\textwidth]{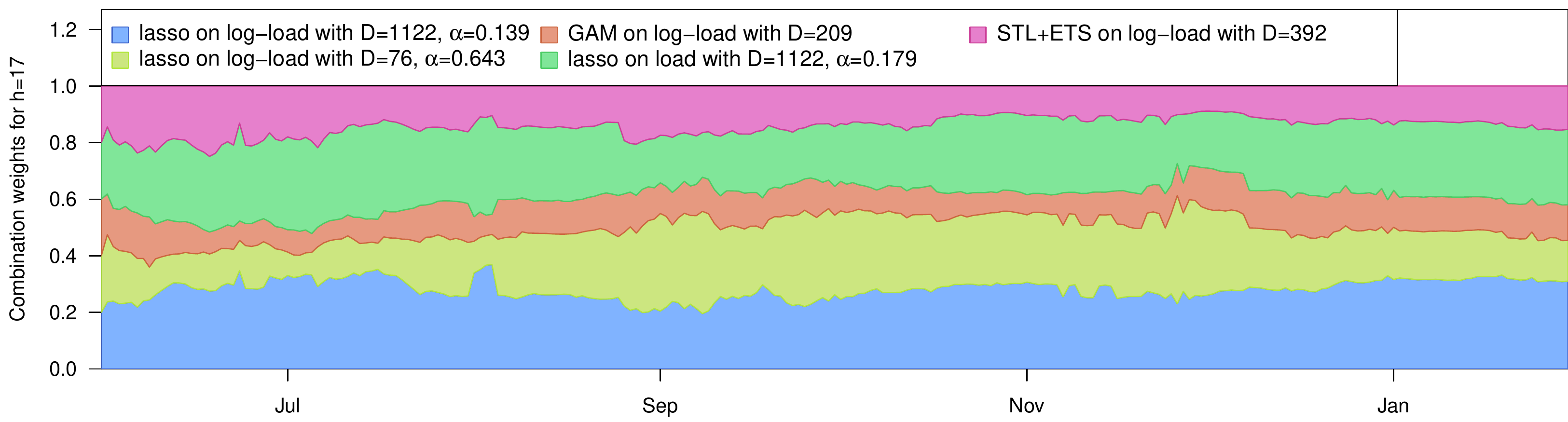}}
\centerline{\includegraphics[width=.999\textwidth]{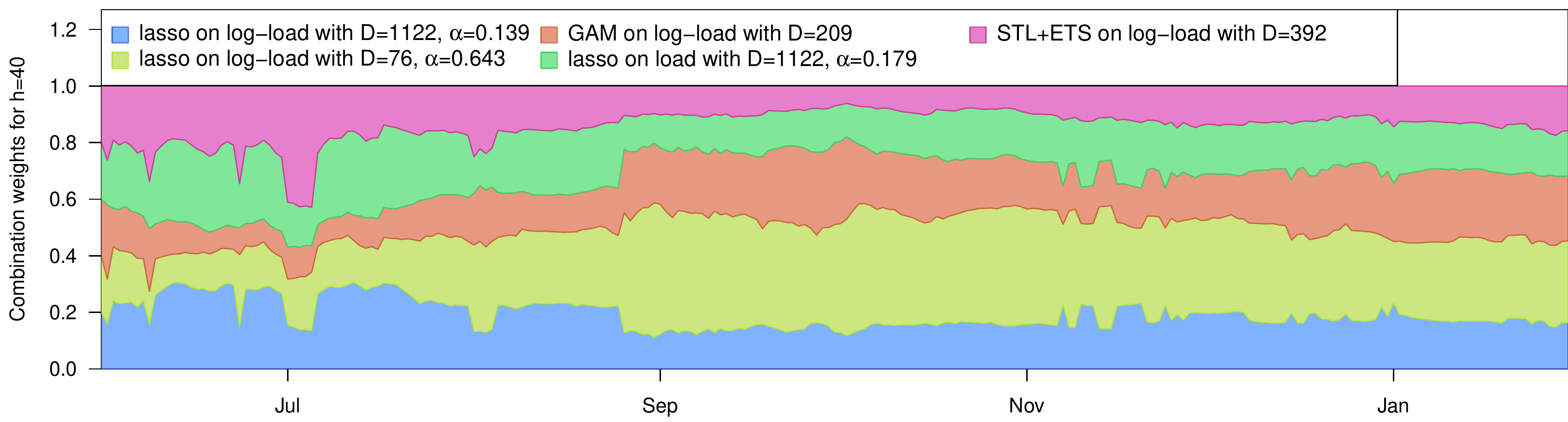}}
\caption{Evolution of combination weights for forecasting horizon $h=17$ and $h=40$, predicting the load at 0:00 and 23:00}
\label{fig_boa_weights}
\end{figure*}

Figure \eqref{fig_boa_weights} visualizes the evolution of the combination weights of the BOA algorithm over time for the forecasting horizons $h=17$ and $h=40$. We observe significant differences, especially the models with short calibration windows (lasso model with $D=76$ and GAM with $D=209$) have more weight for $h=40$.

\begin{figure}[htb!]
\centerline{\includegraphics[width=.499\textwidth]{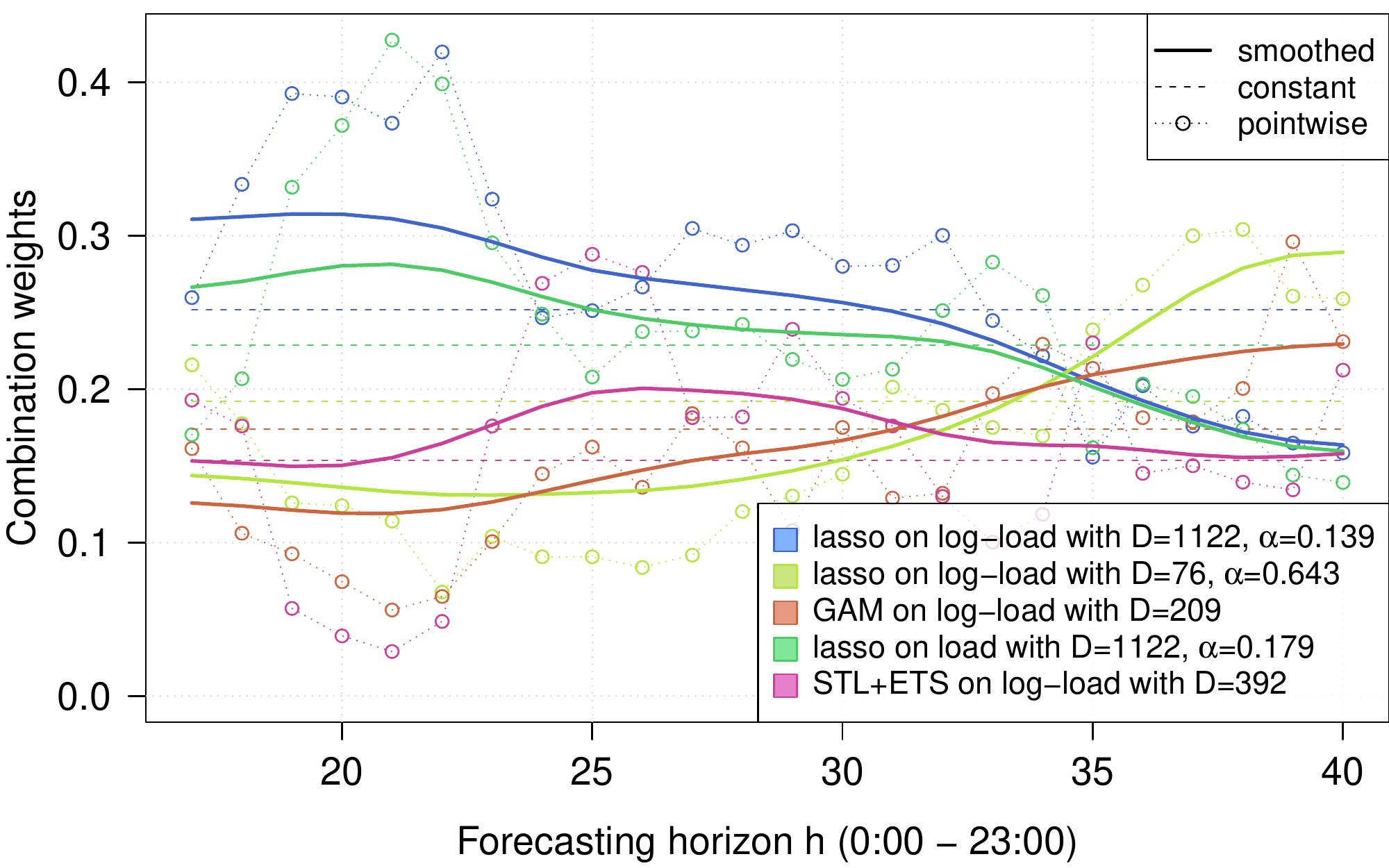}}
\caption{Smoothed combination weights for all forecasting horizons $h=17,\ldots,40$ (0:00 to 23:00) on 1st February 2021. Additionally, we show the limiting constant ($\lambda\to\infty$) and pointwise cases ($\lambda=0$).}
\label{fig_boa_last}
\end{figure}

\begin{figure*}[htb!]
\centerline{\includegraphics[width=.999\textwidth]{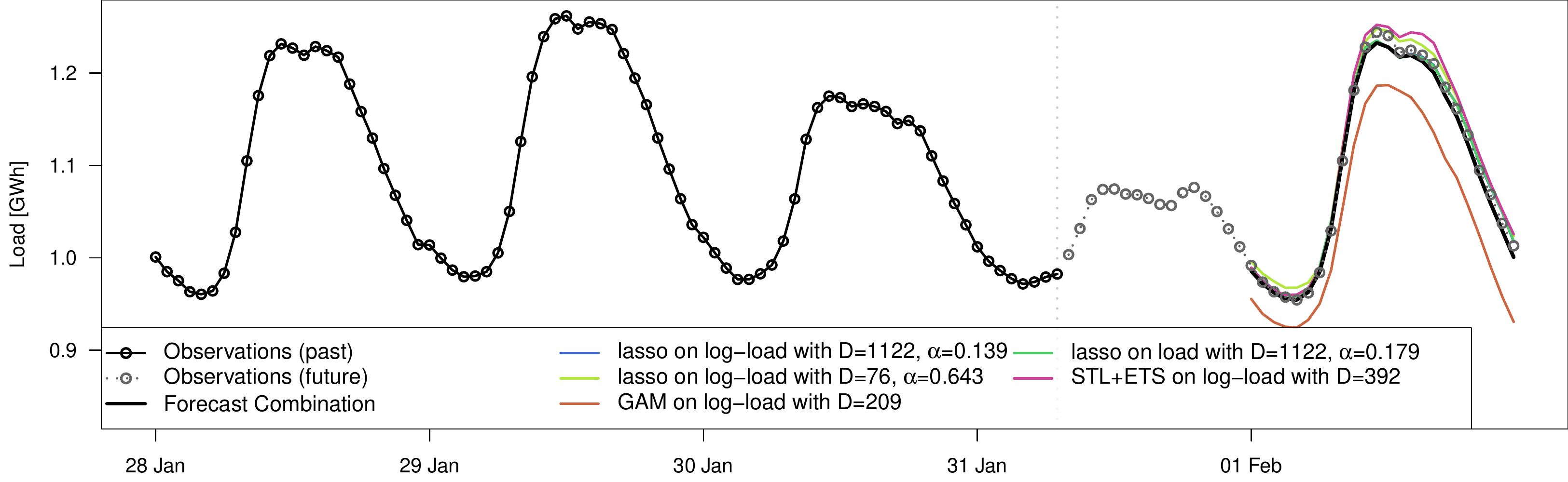}}
\caption{Individual forecasts and forecast combination with observations for the 1st February 2021.}
\label{fig_boa_ts}
\end{figure*}

The same finding can be seen in Figure \eqref{fig_boa_last}. Here, we illustrate the smoothing across the forecasting horizon for the 24 hours in the forecasting horizon. 
We added limiting cases with constant weights ($\lambda\to\infty$) and pointwise optimized weights ($\lambda=0$) to illustrate the effect of smoothing. 
The forecast of the smoothed BOA approach is illustrated in Figure \eqref{fig_boa_ts}. There we see that the GAM model tends to underestimate and the STL+ETS model overestimated the load for the considered forecasting horizon. Thus, they can be regarded as bias correcting models.

\section{Conclusion}

In this manuscript we present one of the winning methods the 
IEEE DataPort Competition on Day-Ahead Electricity Demand Forecasting: Post-COVID Paradigm. It utilizes a sophisticated holiday adjustment procedure, and a novel forecast combination method based on smoothed Bernstein online aggregation (BOA). The approach is flexible and can quickly adopt to new energy system situations.

Obviously, better results may be achieved by more advanced tuning parameter selection design which suffers clearly some optimality.
For instance, some choices on parameter tuning were done ad hoc (e.g. forgetting rate for tuning parameter selection of $\rho=0.01$, validation period of $60$ days) which could be optimized. Furthermore, other BOA extensions as discussed in \cite{berrisch2021crps} like fixed share or regret forgetting could be used as well. Moreover, the pool of individual forecasting models could be enriched as well.
This holds particularly for non-linear models that utilize gradient boosting machines or artificial neural networks. However, the analysis showed that the main features for this short-term load forecasting task are linear, especially the autoregressive and seasonal effects. Hence, no huge improvement should be expected by integrating mentioned models.
%
%
%

\bibliographystyle{IEEEtran}
\bibliography{ref}

\begin{IEEEbiographynophoto}{Florian Ziel}
Florian Ziel is Assistant Professor of Environmental Economics at the House of Energy Markets and Finance at the University of Duisburg-Essen, Germany. He received his M.Sc. in statistics from University College Dublin (Ireland, 2012), his Diplom in mathematics from Dresden University of Technology (Germany, 2013) and his Ph.D. on forecasting in energy markets from the European-University Viadrina in Frankfurt Oder (Germany, 2016). His research interests include data analytics with application to energy markets and systems. He is the first author of various peer-reviewed journal articles, most notably in top-tier IEEE Transactions on Power Systems, Applied Energy, Energy Economics, Renewable and Sustainable Energy Reviews and International Journal of Forecasting.
\end{IEEEbiographynophoto}

\end{document}